# SOME PATTERNS OF SLEEP QUALITY AND DAYLIGHT SAVINGS TIME ACROSS COUNTRIES: A PREDICTIVE AND EXPLORATORY ANALYSIS

Bhanu Sharma [1] and Eugene Pinsky [2]

[1] College of Science, Northeastern University, Boston, MA, USA
[2] Department of Computer Science, Metropolitan College, , Boston University, Boston, MA, USA

## ABSTRACT

*In this study, we analyzed average sleep durations across 61 countries to investigate the impact of Daylight Saving Time (DST) practices. We identified key metrics influencing sleep and employed statistical correlation analysis to explore relationships among these factors. Countries were categorized based on DST observance, and visualizations were generated to compare sleep patterns between DST and non-DST regions. Our findings indicate that, on average, countries that observe DST tend to have better sleep durations compared to those that do not. However, a more nuanced pattern emerged when accounting for latitude: DST- observing countries at lower latitudes reported shorter sleep durations than their non-DST counterparts, whereas at higher latitudes, DST-observing countries demonstrated longer aver- age sleep durations. These results suggest that the effectiveness of DST in improving sleep may be moderated by a country's geographical location.*

## 1. INTRODUCTION

Standard Time was originally established to align midday (12:00 PM) with the Sun's highest point in the sky at a specific reference meridian [1]. The adoption of standardized time began in the 1840s, primarily to coordinate railway schedules, and quickly became necessary for synchronizing social and economic activities across regions [2]. Although based on solar time, time zone boundaries have since been shaped more by political and administrative decisions than by strict geographical or astronomical logic [3].

Many countries observe biannual time changes, transitioning between Standard Time and Day- light Saving Time (DST). The goal of DST is to shift daylight to later in the day during summer months, offering more usable daylight in the evening hours, particularly for populations with work or school obligations during the day. Typically, clocks are moved forward by one hour in spring ("spring forward") and set back by one hour in autumn ("fall back") [4].

The concept of "adjusting" time, however, is not new, Ancient civilizations adjusted their daily schedules according to the sun [5]. It was a more flexible system than DST: days were often divided into 12 hours regardless of daytime, so each daylight hour became progressively longer during the spring and was shorter in the autumn.

The Romans kept time with water clocks that had different scales for different times of the year [6]. For example, on the winter solstice, the third hour from sunrise (hora tertia) started at 09:02 and lasted 44 minutes, whereas during the summer solstice, it started at 06:58 and lasted 75 minutes.

Benjamin Franklin, in a 1784 essay, [7] did suggest that people could save on candles by getting out of bed earlier in the summer. However, this was not a formal proposal for daylight saving time as we know it.

It was Entomologist George Hudson who first proposed modern Daylight Savings Time [8]. This was because his shift-work job gave him leisure time to collect insects, with the result being that he valued after-hours daylight. In 1895, he presented a paper to the Wellington Philosophical Society [9] that proposed a two-hour daylight saving shift forward in October and backward in March. However, the idea was never formally adopted.

Many publications also credited English builder William Willett, who, during a pre-breakfast ride in 1905, observed how many Londoners slept through the sunlit hours of the morning during the summer. He was also an avid golfer who disliked cutting his round short when it got dark [10]. Although many people lobbied for the concept of Daylight Savings, it was not until 1918 that it was formally adopted. The first countries to formally adopt DST were the German Empire and its World War I ally Austria-Hungary in April 1916 as a way of conserving coal during wartime [11, 12].

Britain, most of its allies, and many European neutral countries quickly followed, while Russia waited until a year later, and the US adopted the policy in 1918 as part of the Standard Time Act. The US also re-implemented the policy during World War II.

As of 2025, DST is practiced in approximately 71 countries or autonomous territories, including most of Europe, the United States, Canada, New Zealand, Israel, Chile, and parts of Australia and the Middle East. Some countries, like Morocco, make time changes for religious observances such as Ramadan. The start and end dates of DST vary by country, typically falling in the spring and autumn seasons. Many countries in Africa, Middle East, East Asia and some countries in Latin America historically have never practiced DST.

Whether DST should continue to be practiced has become a subject of ongoing political, social, and scientific debate [13, 14, 15, 16]. The implementation of DST has historically been influenced more by political and historical factors than by geographic considerations. Increasing criticism of biannual clock changes has brought attention to their potential impact on health, productivity, and well-being. Debates focus on whether to abolish DST transitions entirely, and if so, whether permanent DST or permanent Standard Time should be adopted.

In particular, DST has received scrutiny from the sleep science and chronobiology communities [17]. Changes in local clock time can disrupt circadian rhythms, the internal biological systems that align human physiology with natural light-dark cycles. Disruptions to circadian timing, especially due to DST transitions, have been associated with sleep disturbances, reduced performance, and long-term health risks, including cardiovascular, metabolic, and mental health issues [18, 19, 20, 21, 22].

Despite the intensity of the debate, there remains a lack of systematic, data-driven evaluations of DST's actual impact on sleep. This study helps bridge this gap by analyzing sleep duration data from 61 countries, comparing those that observe DST with those that do not. We further investigate how the relationship between DST and sleep varies across geographic latitude, offering new insight into whether DST's impact is consistent or context-dependent. The rest of the paper is organized as follows: In Section 2 we discuss the dataset and its numeric and non-numeric features. In Section 4 we identify correlations between sleep features and demonstrate sleep trends across countries that observe DST versus those that do not. In Section 4.2 We add another dimension to the analysis by examining how latitude affects sleep patterns in DST-observing and non-DST countries. In Section 4.3 we develop and compare classification models aimed at predicting DST implementation based on geophysical variables. Finally, in Section 5 we summarize our findings and analysis to draw broader inferences about DST policy implications, discuss its limitations, and discuss future work.

## 2. DATASET

We used publicly available sleep statistics from Sleep Cycle [23] via their open dataset API at `https://sandman.sleepcycle.com/data`. The original dataset contained average sleep statistics from 61 countries with five core variables: Country, Sleep Quality, Sleep Duration, Wake-up Time, and Bedtime. This is presented in Table 1.

Table 1: Original features in the dataset

| Country | Sleep Quality | Sleep Duration | Snore Duration | Bedtime | Wake-up Time |
|---|---|---|---|---|---|
| Ecuador | 0.7426 | 7.3087 | 3.1281 | 23:29 | 6:46 |
| Singapore | 0.7272 | 7.1064 | 5.4312 | 00:34 | 7:40 |
| Costa Rica | 0.7374 | 7.3553 | 4.3181 | 23:14 | 6:34 |
| … | … | | … | … | … |
| Canada | 0.7742 | 7.6210 | 4.5310 | 23:49 | 7:26 |
| New Zealand | 0.7978 | 7.9865 | 4.8423 | 23:27 | 7:26 |
| Finland | 0.7910 | 7.8737 | 5.0887 | 00:07 | 8:01 |

Sleep Cycle calculates Sleep Quality based on four metrics: time spent in bed, time in deep sleep, motion frequency and intensity, and number of full awakenings. These are combined to generate a personalized sleep quality score, aggregated at the country level.

We augmented the dataset by calculating 6 additional features, including geographic variables (Daylight Savings, hemisphere, and latitude) and seasonal daylight metrics (longest night duration and equinox night duration and their ratios). The dataset includes 36 countries practicing Daylight Saving Time and 25 that do not, enabling comparative analysis across different time regulation systems. This is shown in Table 2.

Table 2: Augmented Features in the Dataset.

| Daylight Savings | Hemisphere | Approximate Latitude | Longest Night | Equinox Night | Longest / Equinox |
|---|---|---|---|---|---|
| 0 | Both | 0° | 12.10 | 12.00 | 1.01 |
| 0 | Northern | 1°N | 12.20 | 12.00 | 1.02 |
| 0 | Northern | 10°N | 12.20 | 12.00 | 1.02 |
| … | … | … | … | … | … |
| 1 | Northern | 60°N | 19.75 | 12.30 | 1.61 |
| 1 | Southern | 41°S | 14.50 | 12.13 | 1.20 |
| 1 | Northern | 64°N | 21.50 | 12.33 | 1.74 |

This dataset spanning multiple continents and latitudes enables investigation of relationships between geographic location, seasonal light variation, daylight saving practices, and sleep characteristics across a global population.

## 3. FEATURE ENGINEERING AND DATA PREPROCESSING

The geographic and daylight variables were derived from published astronomical and geographical databases. Specifically:

1. Latitude values were obtained from standard geographical coordinates for each country's capital city or population center
2. Longest night duration was calculated using astronomical formulas based on the winter solstice at each latitude
3. Equinox night duration was computed for the spring/autumn equinox when day and night are approximately equal Longest/Equinox ratio was calculated to capture the degree of seasonal daylight variation
4. All the correlation matrices were computed using the Pearson correlation coefficient

$$\text{corr}(X, Y) = \frac{\text{Cov}(X, Y)}{\sigma_X \cdot \sigma_Y}$$

Prior to computing the correlation, All features were normalized to achieve zero mean and unit variance, ensuring that variables with different scales contribute equally to the analysis. The normalization transformation applied was:

$$z = \frac{x - \mu}{\sigma}$$

where $x$ is the original value, $\mu$ the mean, and $\sigma$ is the standard deviation.
5. The original dataset contained no missing values for core sleep metrics.
6. Bedtime and wake-up times were converted from HH:MM format to time in minutes (e.g., 23:30 → 1410) to enable numerical analysis. Times after midnight were adjusted for bed- time (e.g., 01:00 → 1500) to maintain temporal chronological order.

## 4. RESULTS

In this article we seek to answer the following questions:

1. What are the most important features or parameters in determining Sleep Quality? How does Sleep Quality vary between DST and non-DST countries? And what effect does DST have on sleep?
2. How does the geographical latitude of a country influence sleep in both DST and non-DST countries?
3. With respect to optimal sleep, is there a way to classify whether a country should employ DST?

### 4.1. Sleep Quality Analysis Across Countries

We start by analyzing statistical metrics for DST and non-DST countries.

Figure 1 presents the correlation matrix for the five primary features influencing sleep across the entire dataset of 61 countries: Sleep Quality, Sleep Duration, Snore Duration, Bedtime, and Wake-up Time. Notably, Sleep Quality exhibits a strong positive correlation with Sleep Duration ($r = 0.97$), indicating that longer sleep durations are associated with higher perceived sleep quality, which aligns with established sleep research [24]. In contrast, Bedtime demonstrates a negative correlation with Sleep Quality, suggesting that earlier bedtimes tend to be linked to improved sleep quality.

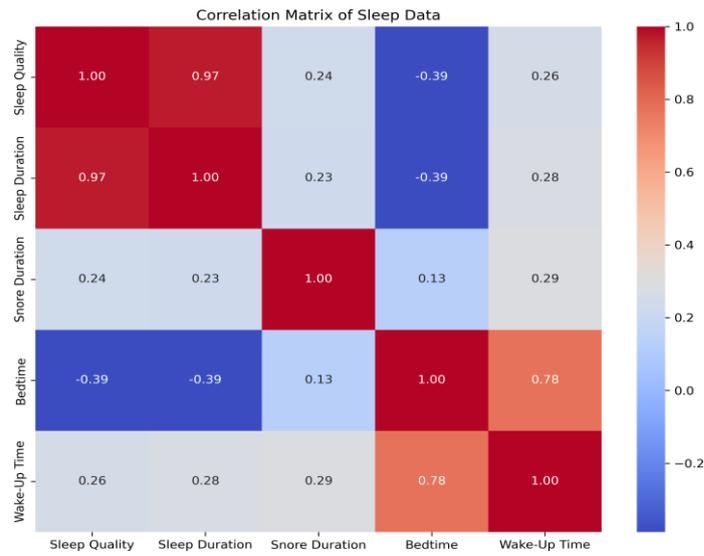

Figure 1: Correlation matrix of sleep parameters across all 61 countries using Pearson correlation coefficient. Color scale: dark red (r = 1.0) indicates perfect positive correlation, white (r = 0) indicates no correlation, and dark blue (r = -1.0) indicates perfect negative correlation. Notable findings: Sleep Quality and Duration show very strong positive correlation (r = 0.97).

The descriptive statistics in Table 3 reveals several notable patterns in sleep behavior across the 61 countries examined. Sleep quality scores averaged 0.75 (with standard deviation $\sigma = 0.029$ hours), indicating generally good sleep quality with remarkably low variability between countries, suggesting consistent sleep satisfaction globally. Sleep duration averaged 7.44 hours (with a standard deviation $\sigma = 0.31$ hours), falling within the recommended 7-9 hour range for adults, with relatively modest cross-country variation.

Temporal sleep patterns showed considerable consistency across nations. The average bedtime was 00:08 ($\sigma = 28$ minutes), clustering tightly around midnight, while wake-up times averaged 7:34 AM ($\sigma = 27$ minutes). This synchronization suggests universal influences on sleep timing, possibly related to work schedules, social norms, or circadian rhythm alignment with daylight cycles.

Table 3: Summary Sleep Statistics for all countries (count = 61)

|  | Sleep Quality | Sleep Duration (hrs) | Snore Duration | Bedtime | Wake-up Time |
|---|---|---|---|---|---|
| Mean | 0.7507 | 7.4352 | 4.4594 | 00:08 | 7:34 |
| Std | 0.0293 | 0.3060 | 0.9263 | 00:28 | 0:27 |
| Min | 0.6766 | 6.7248 | 2.2290 | 22:59 | 6:16 |
| Max | 0.8033 | 7.9865 | 6.3027 | 01:17 | 8:24 |
| 25% | 0.7314 | 7.2327 | 3.9000 | 23:50 | 7:22 |
| Median | 0.7575 | 7.5284 | 4.5604 | 00:05 | 7:37 |
| 75% | 0.7735 | 7.6573 | 4.9779 | 00:29 | 7:55 |

As seen from Table 3, Snoring duration exhibited the greatest variability among measured parameters ($\sigma = 0.93$ hours), with values ranging from 2.23 to 6.30 hours and an average of 4.46 hours. This high variation may reflect differences in sleep monitoring sensitivity, cultural reporting patterns, or genuine physiological differences across populations.

The narrow interquartile ranges for most variables further highlight the global consistency in sleep patterns, with 50% of countries falling within relatively tight bands for sleep quality (0.73-0.77), duration (7.23-7.66 hours), and timing parameters. These findings suggest that de- spite cultural, economic, and geographic diversity, fundamental sleep behaviors show remarkable similarity across countries.

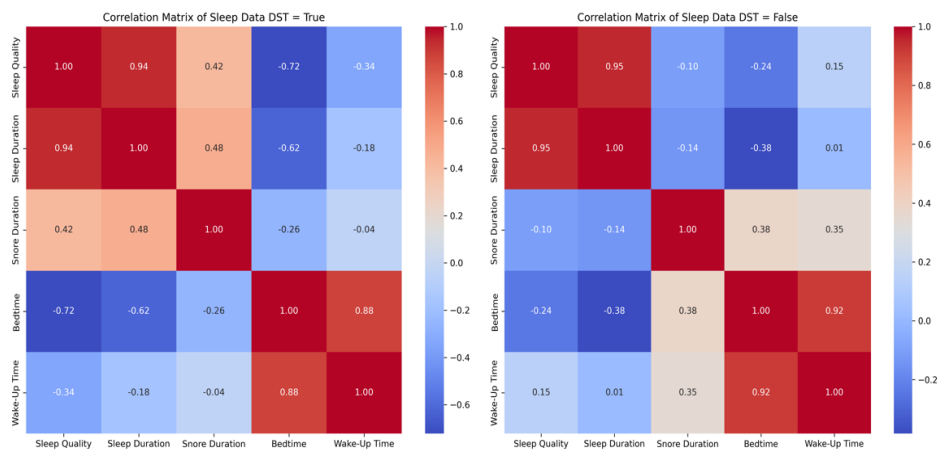

Figure 2: Correlation matrices: (left) countries practicing DST (n = 36), (right) countries not practicing DST (n = 25) using Pearson correlation coefficient. Color scale: dark red (r = 1.0) indicates perfect positive correlation, white (r = 0) indicates no correlation, and dark blue (r = -1.0) indicates perfect negative correlation.

The correlation patterns in Figure 2 reveals striking differences between countries with and without daylight saving time (DST). In DST countries, sleep quality and duration show a very strong positive correlation ($r = 0.94$), while this relationship is similarly strong in non-DST countries ($r = 0.95$). However, the relationship between sleep parameters and timing differs markedly between groups.

Most notably, DST countries exhibit strong negative correlations between sleep quality/duration and bedtime ($r = -0.72$ and $r = -0.62$), suggesting that later bedtimes are associated with poorer sleep outcomes. This pattern is much weaker in non-DST countries ($r = -0.24$ and -0.38).

The correlations in the overall dataset are illustrated in Figure 1 represents a blend of these contrasting patterns, masking the distinct sleep dynamics within each group. The moderate correlations in the overall matrix (bedtime vs. sleep quality: $r = -0.39$) fall between the strong negative relationship in DST countries and the weak relationship in non-DST countries.

Thus, DST implementation creates a sleep environment where timing becomes critical for sleep quality, while non-DST countries demonstrate more timing-independent sleep patterns. This suggests that artificial time shifts from DST may increase sensitivity to circadian misalignment, making sleep timing a more important determinant of sleep health in affected populations [25].

Table's 4 and 5 represent the summary statistics of DST True countries (count = 36) and DST False countries (count = 25). The summary statistics reveal substantial differences in sleep outcomes between DST and non-DST countries, reinforcing the correlation patterns observed earlier. DST countries demonstrate better sleep metrics across all measures: higher sleep quality (0.77 vs 0.73), longer sleep duration (7.63 vs 7.16 hours), and reduced variability in both parameters. The standard deviations are notably smaller for DST countries, indicating more consistent sleep patterns within this group.

Table 4: Summary Sleep Statistics for countries having DST (count = 36)

|         | Sleep Quality | Sleep Duration (hrs) | Snore Duration | Bedtime | Wake-up Time |
|---------|---------------|----------------------|----------------|---------|--------------|
| Mean    | 0.7681        | 7.6262               | 4.6420         | 00:05   | 7:44         |
| Std     | 0.0203        | 0.1917               | 0.6439         | 00:23   | 0:19         |
| Min     | 0.6902        | 6.9320               | 2.3585         | 23:19   | 7:05         |
| Max     | 0.8033        | 7.9865               | 6.0055         | 01:17   | 8:24         |
| 25%     | 0.7605        | 7.5534               | 4.4020         | 23:49   | 7:28         |
| Median  | 0.7717        | 7.6186               | 4.6411         | 00:03   | 7:41         |
| 75%     | 0.7792        | 7.7220               | 4.8662         | 00:18   | 7:57         |

Table 5: Sleep Statistics for countries not practicing DST (count = 25)

|         | Sleep Quality | Sleep Duration (hrs) | Snore Duration | Bedtime | Wake-up Time |
|---------|---------------|----------------------|----------------|---------|--------------|
| Mean    | 0.7257        | 7.1591               | 4.2187         | 00:10   | 7:20         |
| Std     | 0.0219        | 0.2242               | 1.2055         | 00:34   | 0:31         |
| Min     | 0.6766        | 6.7248               | 2.2290         | 22:59   | 6:16         |
| Max     | 0.7737        | 7.7152               | 6.3027         | 01:04   | 8:09         |
| 25%     | 0.7132        | 6.9809               | 3.3196         | 23:51   | 6:58         |
| Median  | 0.7298        | 7.1708               | 4.0320         | 00:08   | 7:28         |
| 75%     | 0.7382        | 7.3087               | 5.0940         | 00:34   | 7:45         |

Non-DST countries show greater sleep variability, particularly in snoring duration ($\sigma = 1.21$ vs 0.64) and timing parameters. Despite similar average bedtimes (00:05 vs 00:10), non-DST countries wake up earlier (7:20 vs 7:44) but achieve shorter sleep duration, suggesting less efficient sleep or different cultural sleep norms.

The timing-quality relationship established in the correlation analysis before is supported by these statistics: DST countries, where timing strongly predicted sleep outcomes, achieve both better sleep quality and more synchronized sleep schedules. This suggests that while DST creates timing sensitivity, it may also promote sleep behaviors that optimize circadian alignment, resulting in overall better sleep health compared to non-DST populations.

Analysis of correlation pattern deviations reveals that DST countries exhibit markedly distinct sleep relationships compared to the global dataset. Despite DST countries comprising the majority of the sample (36 of 61 countries), their correlation patterns showed a greater distance from the overall correlations (norm distance = 1.48) compared to non-DST countries (norm distance = 0.95). This counterintuitive finding suggests that DST implementation creates more extreme and specialized sleep correlation structures that significantly deviate from global norms.

Typically, the larger subgroup would be expected to drive the overall patterns, resulting in smaller deviations. However, the pronounced distance for DST countries indicates that daylight saving time fundamentally restructures sleep parameter relationships, creating heightened sensitivity to timing factors that diverges markedly from natural sleep patterns. Non-DST countries, despite representing the smaller subset, maintain correlation patterns closer to a baseline sleep ecology, suggesting their sleep relationships may reflect more universal, biologically driven associations uninfluenced by artificial time shifts.

Table 6: Correlation matrix means and standard deviations across categories

| Feature | Metric | Overall | DST True | DST False |
|---|---|---|---|---|
| Sleep Quality | mean | 0.4149 | 0.2578 | 0.3516 |
| | std | 0.5207 | 0.6864 | 0.5247 |
| Sleep Duration | mean | 0.4192 | 0.3222 | 0.2883 |
| | std | 0.5195 | 0.6331 | 0.5757 |
| Snore Duration | mean | 0.3779 | 0.3186 | 0.2980 |
| | std | 0.3154 | 0.4385 | 0.4134 |
| Bedtime | mean | 0.2255 | 0.0547 | 0.3359 |
| | std | 0.5774 | 0.7395 | 0.5707 |
| Wake-Up Time | mean | 0.5201 | 0.2637 | 0.4855 |
| | std | 0.3088 | 0.5611 | 0.4026 |

Table 6 provides the mean and standard deviations exhibited by the Sleep Features across categories: Overall (full dataset n = 61), Countries following DST ($n = 36$) and Countries not following DST ($n = 25$).

The distribution plots in Figure 3 clearly demonstrate the systematic sleep advantage in DST countries. For sleep duration, DST countries (blue) predominantly cluster above the global median of 7.53 hours, with most values ranging from 7.5 to 8.0 hours, while non-DST countries (red) are heavily concentrated below the median, primarily between 6.9 and 7.4 hours.

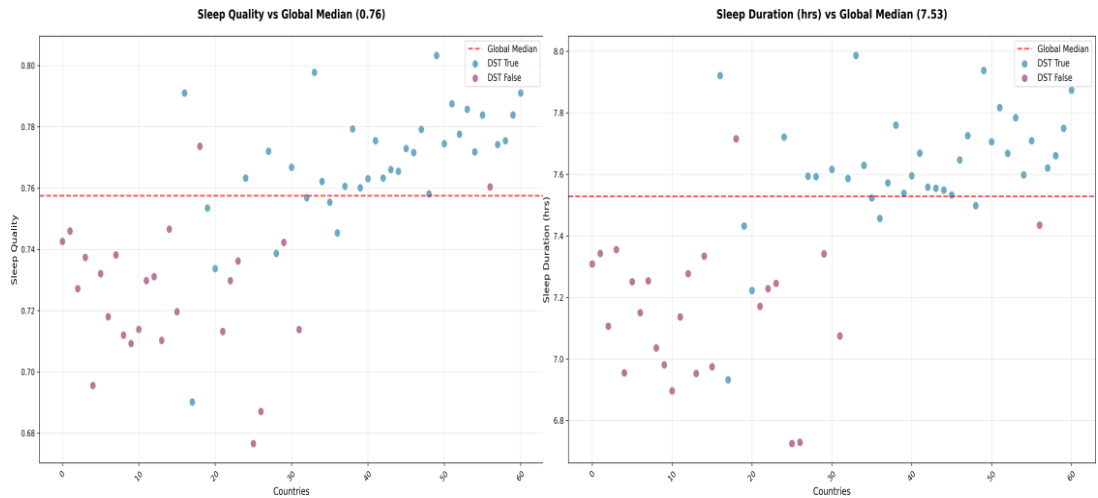

Figure 3: Country wise Sleep Quality and Sleep Duration vs Global Median

The sleep quality pattern mirrors this distribution: DST countries consistently exceed the global median of 0.76, with most countries achieving quality scores between 0.76 and 0.80. In contrast, non-DST countries show a bimodal distribution with the majority falling below the median (0.68-0.75 range), though a few countries perform comparably to DST nations.

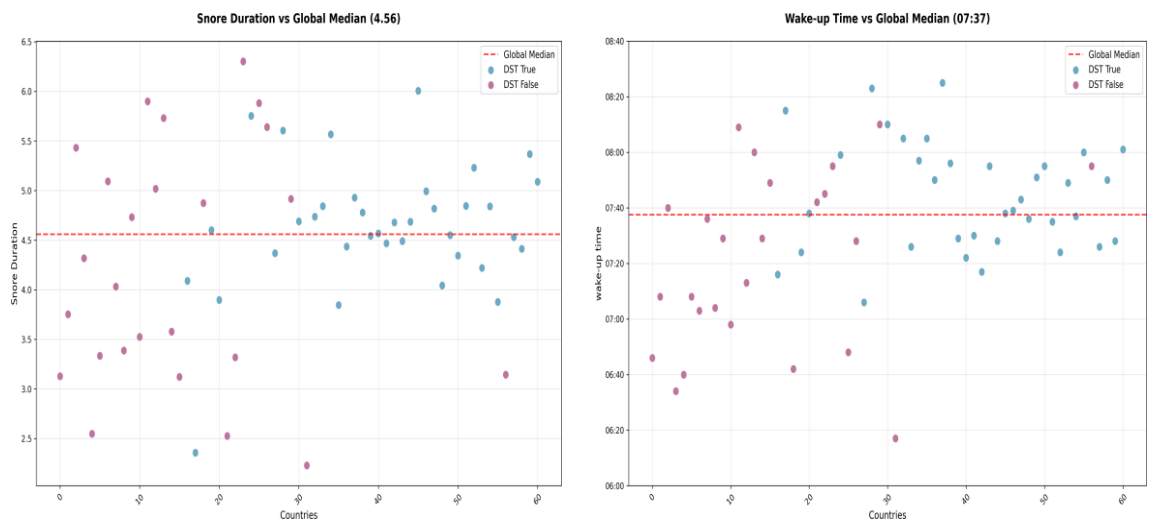

Figure 4: Country wise Snore Duration and Bed Time vs Global Median

According to Figure 4, the snoring duration and wake-up time distributions reveal more com- plex patterns compared to sleep quality and duration. For snoring, both DST and non-DST countries show considerable scatter around the global median of 4.56 hours, with no clear systematic advantage for either group. Non-DST countries display slightly higher variability, including some extreme values above 6 hours, while DST countries cluster more tightly around the median.

Wake-up time patterns show a notable divergence: DST countries predominantly wake up later than the global median of 7:37 AM, with most countries clustering between 7:40 and 8:25 AM. In contrast, non-DST countries exhibit a bimodal distribution, with many countries waking substantially earlier (6:15-7:15 AM) and others aligning closer to the global median. This suggests that DST implementation may promote later wake times, potentially allowing for the longer sleep durations observed earlier, while non-DST countries show more diverse cultural or occupational timing patterns that result in earlier rising times but shorter overall sleep.

## 4.2. Latitude Considerations

Until now, we have established how sleep is affected by Daylight Saving Time (DST) by analyzing features such as sleep quality, sleep duration, snore duration, wake-up time, and bedtime.

The following section presents interesting insights and patterns that emerge from analyzing how sleep is affected by DST in relation to the geographical locations of different countries.

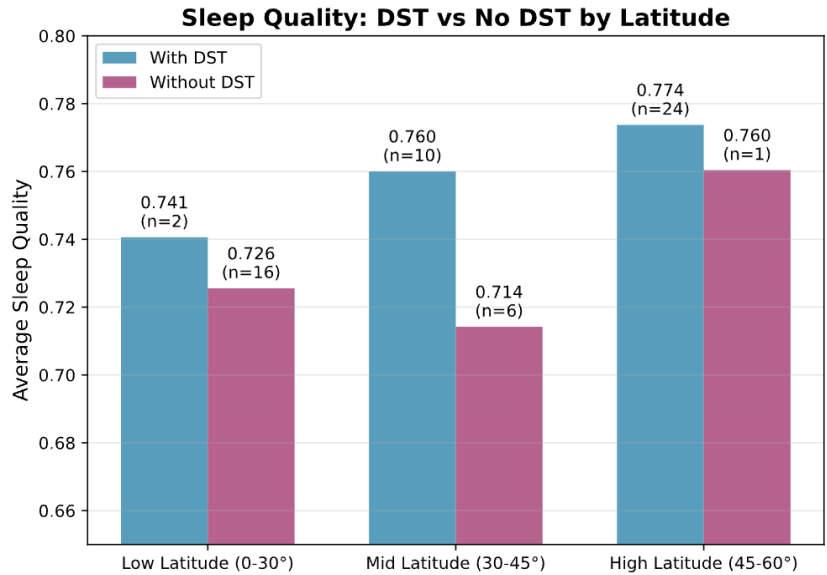

Figure 5: Sleep Quality According to Latitude

Figure 5 reveals latitude-dependent effects of Daylight Saving Time (DST) on sleep quality. At low latitudes (0-30°), countries implementing DST show marginally higher average sleep quality (0.741) compared to non-DST countries (0.726), though this difference is relatively small and based on limited sample sizes (n=2 for DST, n=16 for non-DST countries).

The most pronounced negative effect of DST emerges at mid-latitudes (30-45°), where DST-implementing countries demonstrate substantially higher sleep quality scores (0.760) than their non-DST counterparts (0.714). This 0.046-point difference represents the largest gap observed across all latitude bands, suggesting that DST may have the most disruptive impact on sleep in temperate regions where seasonal daylight variation is moderate.

At high latitudes (45-60°), the sleep quality difference between DST and non-DST countries appears negligible (0.774 vs 0.760). However, due to limitations in available data, the high latitude region is not truly representative, with an extremely small sample size for non-DST countries (n=1). It is expected that at even higher latitudes, the difference in sleep quality between DST and non-DST countries would be more pronounced, as populations in these regions experience extreme seasonal variations in daylight that could exacerbate the circadian disruption caused by artificial time changes.

As shown in Figure 6, we observe the following:

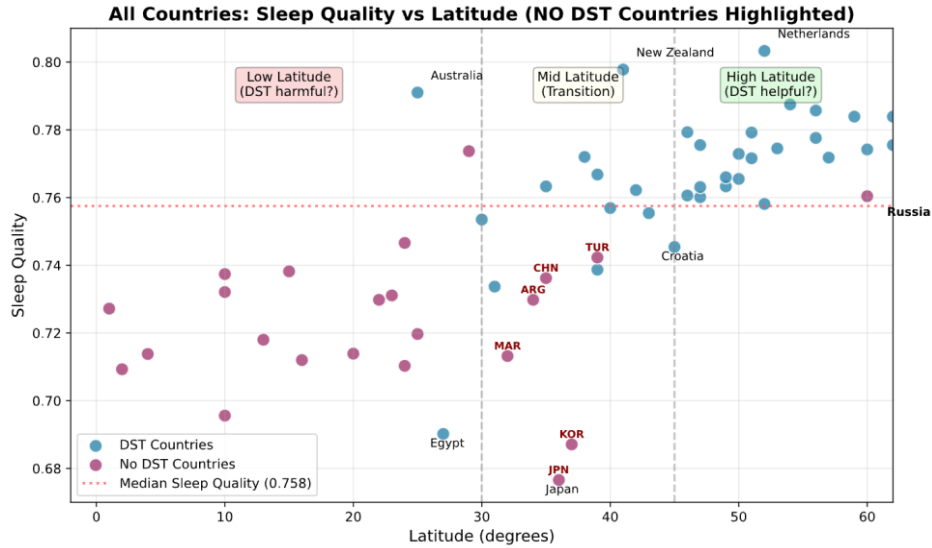

Figure 6: Effect of Latitude on DST

- Low Latitudes (0-30°): the effect of DST appears to be largely non detrimental to sleep quality. This is evidenced by the high variability observed among DST-implementing countries in this region, where Australia achieves exceptional sleep quality (approximately 0.79) while Egypt demonstrates poor sleep quality (approximately 0.69). Both countries represent clear outliers from the non-DST cluster, but in opposite directions. This substantial variation be- tween DST countries suggests that the one-hour time shift has minimal systematic impact on sleep in equatorial regions, where seasonal daylight variation is naturally limited. The observed differences between Australia and Egypt are likely attributable to other socioeconomic, cultural, or environmental factors rather than DST implementation itself.
- Mid Latitudes (30-45°): Emergence of DST Benefits A clearer pattern emerges at mid-latitudes, where countries implementing DST consistently demonstrate better sleep quality compared to their non-DST counterparts. DST countries in this range cluster predominantly above the median sleep quality threshold, while non-DST countries like Turkey, China, Argentina, Morocco, Korea, and Japan fall below or near the median. This suggests that in temperate regions with moderate seasonal daylight variation, the artificial extension of evening daylight through DST may provide measurable sleep quality benefits.
- High Latitudes (45-60°): Pronounced DST Advantage The difference becomes even more pronounced at higher latitudes, where DST-implementing countries show remarkably consistent and superior sleep quality performance. Nearly all DST countries in this range achieve sleep quality scores well above the median (0.76-0.81), forming a tight cluster that contrasts sharply with the single non-DST country (Russia) at the median level. This pronounced advantage suggests that DST may be particularly beneficial in regions experiencing extreme seasonal light variations, potentially by better synchronizing social schedules with the dramatic shifts in natural photoperiods characteristic of higher latitudes.

Overall, this latitude-based analysis indicates that the effectiveness and impact of DST on sleep quality are closely linked to a country's geographic position, with minimal effects in low latitudes and progressively greater benefits observed in mid to high latitudes due to their varying exposure to natural daylight across seasons.

It is important to note a limitation in our dataset: only 1 non-DST country (Russia) is represented at high latitudes (45-60°), compared to 24 DST countries in the same range. This imbalance limits our ability to make comparative claims about DST effectiveness at high latitudes. Future research with more balanced representation of non-DST countries at high latitudes is essential and required.

### 4.3. Machine Learning Results

Next, we develop and compare classification models aimed at predicting DST implementation based on geographical variables

To empirically evaluate whether daylight saving time (DST) is more appropriate for countries at higher latitudes, we developed a set of binary classification models aimed at predicting DST based on geophysical variables. The dataset consisted of 61 countries, each labeled according to whether they currently observe DST. Two features were used: latitude and the ratio of the longest night to the equinox night, which captures seasonal daylight variation.

#### 4.3.1. Methodology and Model Selection

**4.3.1.1. Feature Scaling:** All input features were standardized using StandardScaler to transform data to zero mean and unit variance, preventing features with larger scales from dominating the learning process.

**4.3.1.2. Model Selection and Hyperparameters:** Four classification algorithms were evaluated with the following configurations:

4.3.1.2.1. **Random Forest:** n_estimators=6, max_depth=None, n_jobs=-1 (ensemble of 6 decision trees with unlimited depth).

4.3.1.2.2. **K-Nearest Neighbors (KNN):** n_neighbors=4, weights='uniform', p=2 (4 nearest neighbors with equal weights using Euclidean distance).

4.3.1.2.3. **Gaussian Naive Bayes:** Default parameters (assumes features follow a normal distribution).

4.3.1.2.4. **Logistic Regression:** penalty='l2', C=1.0, solver='lbfgs', max_iter=1000 (L2 regularization with regularization strength $C = 1.0$).

**4.3.1.3. Cross-Validation Strategy:** To ensure robust performance estimates given the limited sample size ($n = 61$), we employed 10-fold cross-validation using sklearn.model_selection.KFold.

Despite the limited sample size, we trained and tested four supervised classifiers: k-nearest neighbors (KNN), random forest, logistic regression, and Gaussian Naive Bayes using 10-fold cross- validation to assess performance robustness. This is shown in Figure 7.

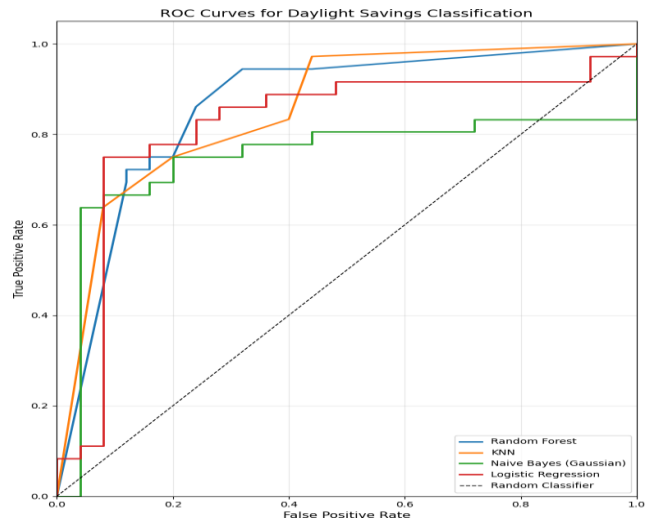

Figure 7: Receiver Operating Characteristic (ROC) curves for four classification models predicting DST implementation.

The diagonal line represents random classification (AUC=0.5). Models performing above this line demonstrate predictive capability, with Random Forest achieving the highest AUC of 0.863, indicating strong discriminative power between DST and non-DST countries based on geographic features.

All models achieved a mean accuracy exceeding 78%, with ROC curves plotted for each classifier and corresponding AUC (Area Under the Curve) scores calculated to quantify discriminative ability as shown in Table 7.

Table 7: Cross-validated performance metrics (10 folds) for DST classification models

| Classifier | AUC | Accuracy | Precision | Recall | $f_1$ Score |
|---|---|---|---|---|---|
| Random Forest | 0.863 | 0.802 | 0.860 | 0.843 | 0.830 |
| K-Nearest Neighbors | 0.855 | 0.783 | 0.885 | 0.793 | 0.811 |
| Logistic Regression | 0.820 | 0.819 | 0.840 | 0.935 | 0.866 |
| Naive Bayes (Gaussian) | 0.749 | 0.783 | 0.832 | 0.868 | 0.827 |

These results suggest that simple geophysical parameters may hold predictive value in determining the practical applicability of DST, especially in distinguishing between higher and lower latitude regions.

## 5. DISCUSSION AND CONCLUDING REMARKS

These findings suggest that while DST is associated with better sleep on a global scale, latitude plays a crucial moderating role. At lower latitudes, the effect of DST appears minimal, likely due to relatively stable daylight hours year-round.

However, at higher latitudes, where day length varies more dramatically across seasons, DST may help align social and biological clocks, enforcing a kind of 'discipline' and promoting more consistent sleep patterns. This supports the argument that the decision to implement DST should consider geographical factors such as latitude rather than being driven solely by historical or political traditions and conventions [26].

Further, we investigated whether geophysical factors, specifically latitude and the ratio of longest night to equinox night, can predict the usage of Daylight Saving Time (DST) across countries. Using a dataset of 61 countries labeled by their current DST observance, we trained four standard classifiers: K-nearest neighbors (KNN), random forests, logistic regression, and Gaussian Naive Bayes. Despite the limited dataset size and feature set, all models achieved mean accuracies above 80% under 10-fold cross-validations. ROC curve analysis and Area Under the Curve (AUC) metrics indicated strong separability between countries that do and do not observe DST as well.

These results suggest that latitude and seasonal daylight variation are not only correlated with DST usage but may be informative for predicting DST relevance. This provides quantitative sup- port for the argument that geophysical realities rather than solely political or historical precedent could better inform DST policy.

However, several limitations must be acknowledged. The small sample size (n = 61) limits the generalizability of our conclusions. Additionally, while high classification accuracy suggests predictive strength, it does not imply causation. DST policies are ultimately the product of complex political, cultural, and economic decisions and may not always reflect optimal alignment with natural daylight patterns. Moreover, the labels in our dataset reflect current political choices, which may themselves be inconsistent or misaligned with health and energy considerations.

In conclusion, this study provides preliminary empirical evidence that DST observance tends to align with geophysical patterns, particularly higher latitudes and more extreme seasonal daylight variation. These findings merit further investigation with larger datasets and complementary variables (e.g., economic activity, sleep data, or health outcomes). Policymakers may benefit from incorporating such data-driven insights when evaluating the merits of DST in different regions.

## DECLARATIONS

### AUTHOR CONTRIBUTIONS

All authors contributed equally

### DATA AVAILABILITY

The data supporting the reported results in this study are available on GitHub at `https://github.com/anongit21/Sleep-Patterns`

### ACKNOWLEDGEMENTS

All aspects of the study, including design, data collection, analysis, and interpretation, were carried out using the resources available within the authors' institution(s).

### CONFLICT OF INTEREST

There are no conflicts of interest regarding the publication of this paper.